\documentclass[letterpaper, 10 pt, conference]{ieeeconf} 
\def\IEEEtitletopspace{18pt}
\IEEEoverridecommandlockouts
\usepackage{cite}
\usepackage{amsmath,amssymb,amsfonts}

\usepackage{algpseudocode}
\usepackage{algorithm}
\usepackage{graphicx}
\usepackage{textcomp}
\usepackage{xcolor,colortbl} 
\usepackage{subcaption}
\usepackage{geometry}
\usepackage[nospace]{varioref}
\usepackage{hyperref}

\hypersetup{
    unicode=false,          
    pdftoolbar=true,        
    pdfmenubar=true,        
    pdffitwindow=false,     
    pdftitle={PedAnalyze - Pedestrian Behavior Annotator},    
    pdfauthor={Taorui Huang},     
    pdfsubject={Research Paper},   
    pdfcreator={Taorui Huang},   
    pdfproducer={},  
    pdfnewwindow=true,      
    colorlinks=true,       
    linkcolor=[RGB]{0, 100, 255},          
    citecolor=[RGB]{0, 100, 255},        
    filecolor=magenta,      
    urlcolor=[RGB]{2, 120, 210}          
}

\def\BibTeX{{\rm B\kern-.05em{\sc i\kern-.025em b}\kern-.08em
    T\kern-.1667em\lower.7ex\hbox{E}\kern-.125emX}}
    
\begin{document}
\arrayrulecolor[RGB]{190, 180, 180}   
{\renewcommand{\arraystretch}{1.2}
\title{
Pedestrian Archetypes Extension - More Pedestrian Models for Autonomous Vehicle Safety Testing
}

\author{ Taorui Huang$^{1*}$, Namita Gaidhani$^{2*}$, Ritvik Bansal$^{3*}$, S M Jubaer$^{4*}$, \\ Regina Lim$^{5*}$,
Rhett Zhao$^{6*}$, Gavin Rafael Selin$^{7*}$, Sunnie Deng Gao$^{8*}$, Hasnain N Syed$^{9*}$
%
    \thanks{$^{1}$Stanford University, Stanford, CA, USA.}%
    \thanks{$^{2}$University of California San Diego, San Diego, CA, USA.}%
    \thanks{$^{3}$University of Washington, Seattle, WA, USA.}%
    \thanks{
    {$^{1}$\tt\small taoruih@stanford.edu},
    {$^{2}$\tt\small ngaidhani@ucsd.edu}}
    \thanks{
    {$^{3}$\tt\small ritvikb@uw.edu}}
    \thanks{
    {$^{*}$\tt\small equal contributions.}}
}

\maketitle

\begin{abstract}
In our prior work, Pedestrian Archetypes \cite{archetypes-part1}, we defined pedestrian archetypes as collections of behaviors that uniquely identify a specific type of pedestrian. The first paper proposed 12 pedestrian archetypes, including the Wanderer, Drunk, Distracted, Flash, Indecisive, Blind, Flock, Jaywalker, Elderly, Kid, Eventful, and Parked Pedestrian. These archetypes were introduced to move beyond single behavior labels and provide a more natural way to describe how dangerous pedestrians actually behave progressively in real-world traffic scenarios. However, upon further annotation of YouTube dash-cam videos, we identified 7 additional pedestrian archetypes with observable and significant behavioral differences from the previously proposed ones. These new archetypes capture pedestrian behavior patterns that could not be fully explained by the original taxonomy. In this pre-print, we introduce each new archetype, define its essential and optional behaviors, explain how it differs from previously proposed archetypes, and provide video-frame evidence showing the archetype in action.
\end{abstract}


\section{Introduction \& Motivation}
Pedestrian behavior modeling for autonomous vehicle (AV) safety has traditionally focused on predicting whether a pedestrian will cross, where they will cross, and their expected trajectory. While these factors are important, they do not fully capture the complexity of dangerous pedestrian behavior. In many real-world scenarios, risk arises not from a single action but from a recurring pattern of behaviors that collectively make pedestrian intent and future actions difficult to predict.

Our prior work, PedAnalyze~\cite{PedAnalyze}, introduced a pedestrian behavior ontology consisting of standardized behavior tags and definitions. These tags provide a consistent framework for annotating observable pedestrian actions, such as crossing outside a crosswalk, retreating, flinching, ignoring traffic, entering traffic unexpectedly, or walking along a travel lane. This ontology enables systematic and reproducible behavior annotation across diverse scenarios.

However, behavior tags alone provide only a low-level description of observed actions. While they capture what a pedestrian is doing, they do not necessarily characterize the broader behavioral pattern. For example, the same retreat behavior may arise from indecision, distraction, or a reaction to an external event. Likewise, crossing outside a crosswalk may be associated with multiple pedestrian types, each exhibiting different behavioral tendencies. Behavior tags describe observable evidence; archetypes describe the higher-level pedestrian model.

Pedestrian archetypes address this limitation by grouping related behaviors into coherent behavioral models. Each archetype is defined by a set of essential and optional behaviors that frequently occur together, providing a more complete representation of pedestrian behavior than individual tags alone. This abstraction is particularly valuable for AV safety testing, where anticipating likely future actions is often as important as recognizing current behavior.

In our original archetypes paper~\cite{archetypes-part1}, we proposed 12 pedestrian archetypes as an initial taxonomy of dangerous pedestrian behavior. Continued annotation of YouTube dash-cam videos, however, revealed additional recurring behavior patterns that were visually observable and behaviorally distinct from the original archetypes. These findings suggest that the dangerous pedestrian test space extends beyond the initial taxonomy.

This preprint expands the pedestrian archetype framework by introducing 7 additional archetypes. For each archetype, we provide a formal definition, representative video evidence, comparisons with related archetypes where appropriate, and a characterization of its essential and optional behavior tags derived from the current annotated dataset.

\section{Methodology} \label{methodology}
We identified the new archetypes through continued annotation of YouTube dash-cam footage containing risky, near-miss, and collision pedestrian-vehicle interactions.

The annotation process consisted of four steps. First, pedestrian behaviors were annotated using the PedAnalyze behavior ontology~\cite{PedAnalyze}, which captures observable actions such as crossing, retreating, running, pausing, ignoring traffic, changing direction, and hesitating. Second, annotated scenarios were reviewed to identify recurring behavior combinations and compared against the 12 original archetypes to determine whether they represented previously unmodeled patterns. Third, representative video evidence was selected for each new archetype, and key frames were extracted to illustrate the behavioral sequence and support the archetype definition. Fourth, essential and optional behavior tags were derived from the annotated dataset.

A behavior was classified as essential if it appeared in at least 40\% of examples for a given archetype and optional if it appeared in 10–39\% of examples. Behaviors occurring in fewer than 10\% of examples were generally excluded unless they were qualitatively significant. This threshold-based approach provides a data-driven characterization of each archetype by distinguishing core behaviors from those that are commonly associated but not required.

\section{Archetypes} \label{archetypes}

This section quickly summarizes existing archetypes~\cite{archetypes-part1} and introduces the 7 new pedestrian archetypes identified through continued dash-cam video annotation. For any unclear pedestrian behaviors, please reference our documentation definitions at Appendix \ref{appendix-pedestrian-behaviors}.

\begin{enumerate}
    \item \textit{The Wanderer}: wanders along driving lanes and often ignores traffic with unpredictable intention.
    \item \textit{The Drunk}: is impaired by alcohol or drugs, producing unpredictable gait, balance, and decisions.
    \item \textit{The Distracted}: is rational but not fixated on traffic (ex. on the phone), reducing situational awareness.
    \item \textit{The Flash}: sprints through traffic with urgency and little regard for safety.
    \item \textit{The Indecisive}: hesitates or repeatedly changes crossing intent, propagating uncertainty and increasing risk.
    \item \textit{The Blind}: intentionally ignores or fails to notice vehicles and traffic signals.
    \item \textit{The Flock}: is a group that typically moves together. They become dangerous when members disperse or re-group mid-crossing.
    \item \textit{The Jaywalker}: crosses at non-designated locations (i.e. mid-block) and on red, often choosing the shortest path.
    \item \textit{The Elderly}: exhibits slow decision-making, takes longer than expected to cross due to slower walking speeds.
    \item \textit{The Kid}: inexperienced, impulsive, and difficult to see.
    \item \textit{The Eventful}: becomes dangerous due to external, involuntary events (i.e. trips, dropped items, or pets).
    \item \textit{The Parked Pedestrian}: interacts with parked vehicles (ex. loading, getting in/out), often occluded by traffic.
\end{enumerate}

\subsection{The Con Artist}\label{sec:the-fraud}

\begin{figure}[bt!]
    \centering
    \begin{subfigure}[b]{.49\linewidth}
      \centering
      \includegraphics[width=.99\linewidth]{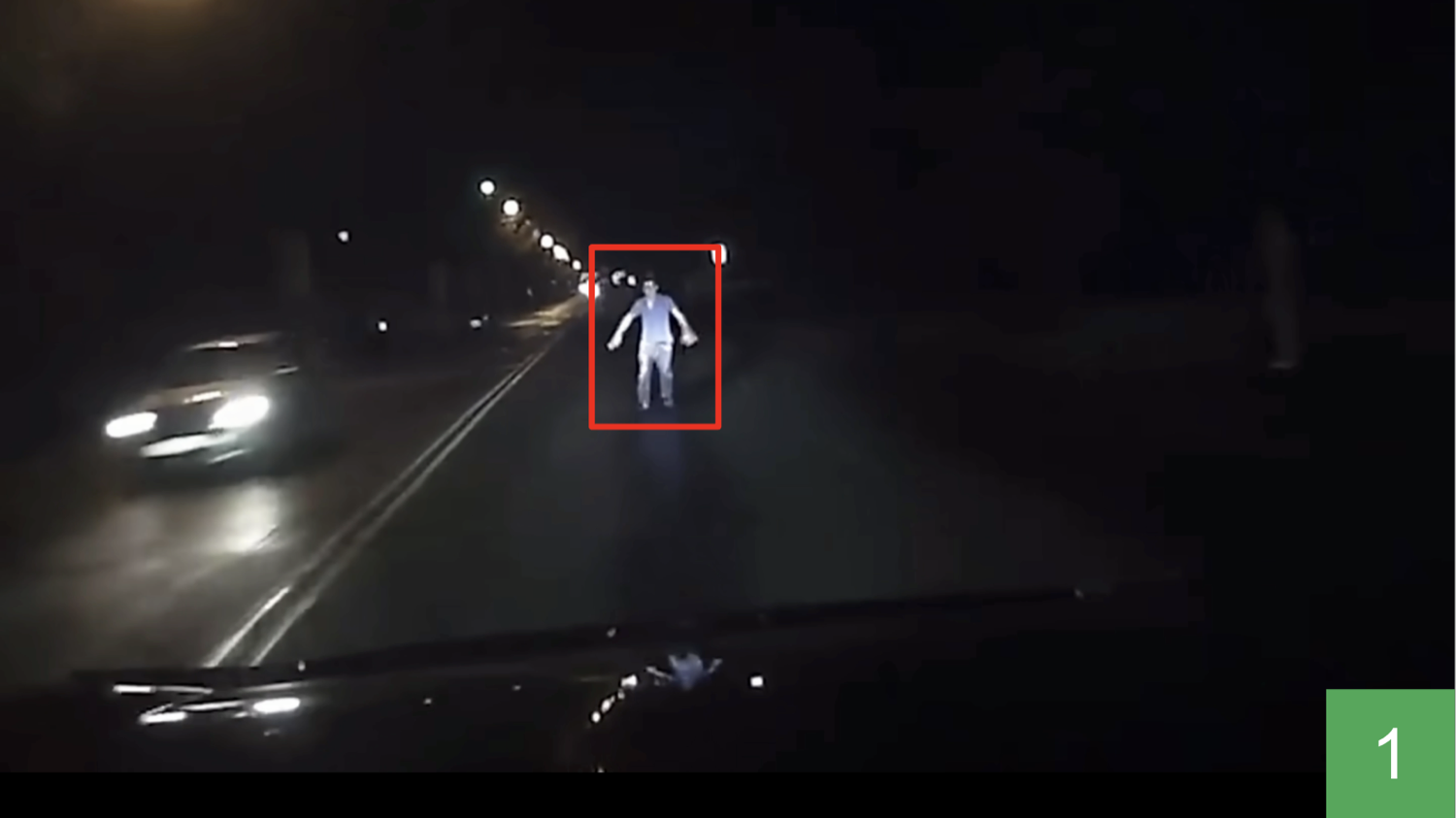}
      \caption{Stands in middle of lane}
      \label{fig:archetype-fraud-1}
    \end{subfigure}
    \hfill
    \begin{subfigure}[b]{.49\linewidth}
      \centering
      \includegraphics[width=.99\linewidth]{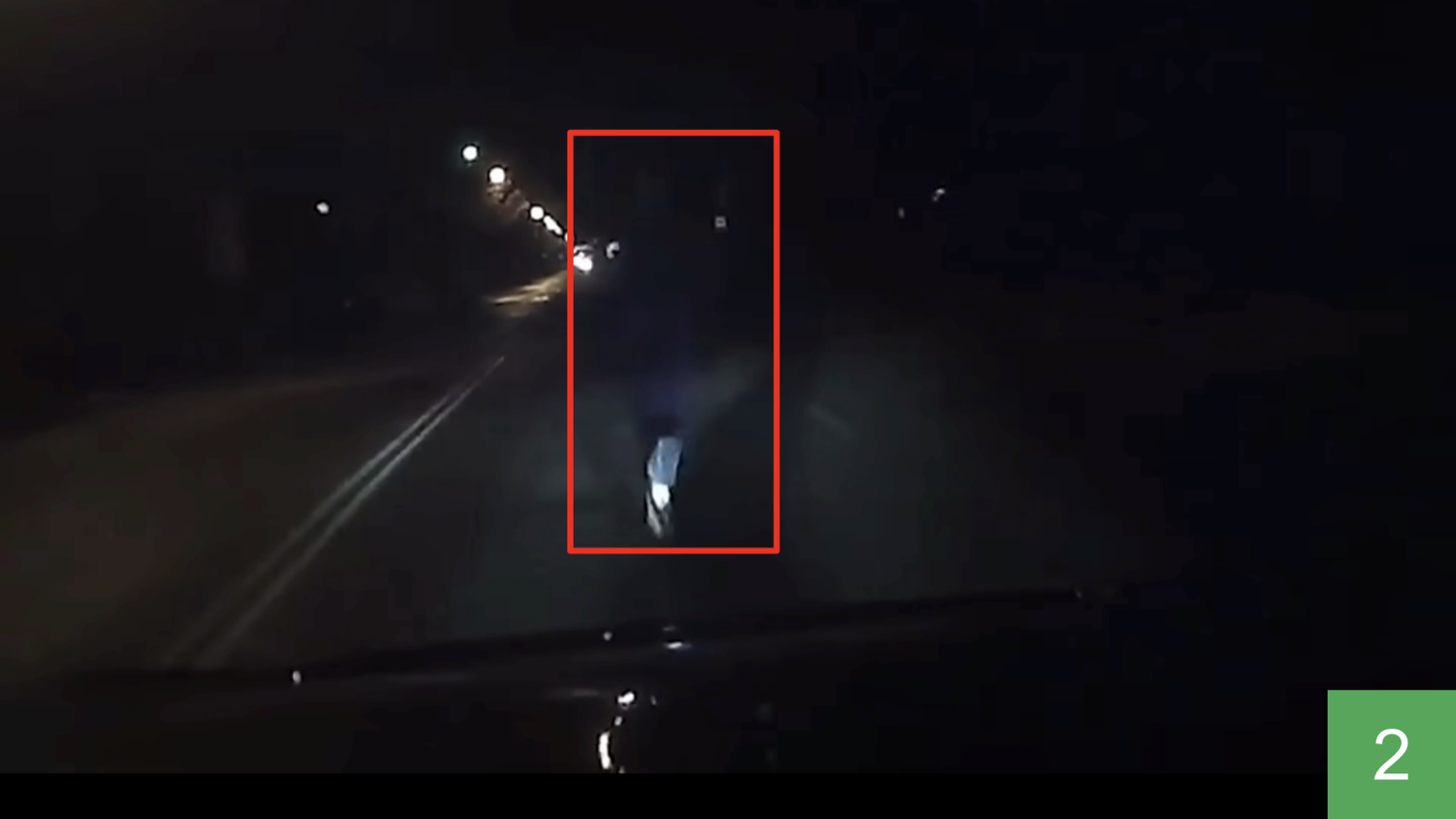}
      \caption{Intentional jump}
      \label{fig:archetype-fraud-2}
    \end{subfigure}
    
    \begin{subfigure}[b]{.49\linewidth}
      \centering
      \includegraphics[width=.99\linewidth]{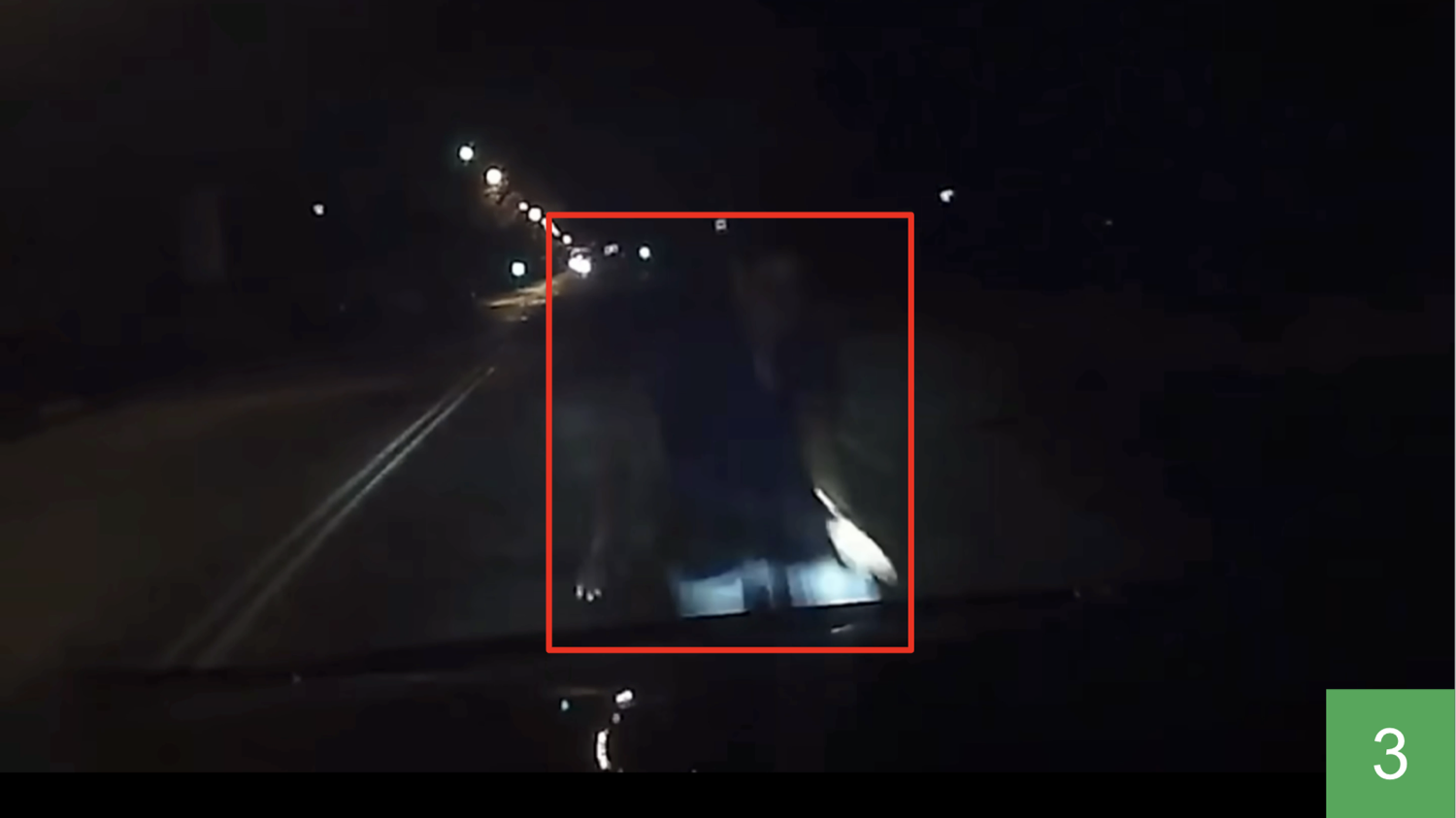}
      \caption{Contact made, drops to ground}
      \label{fig:archetype-fraud-3}
    \end{subfigure}
    \hfill
    \begin{subfigure}[b]{.49\linewidth}
      \centering
      \includegraphics[width=.99\linewidth]{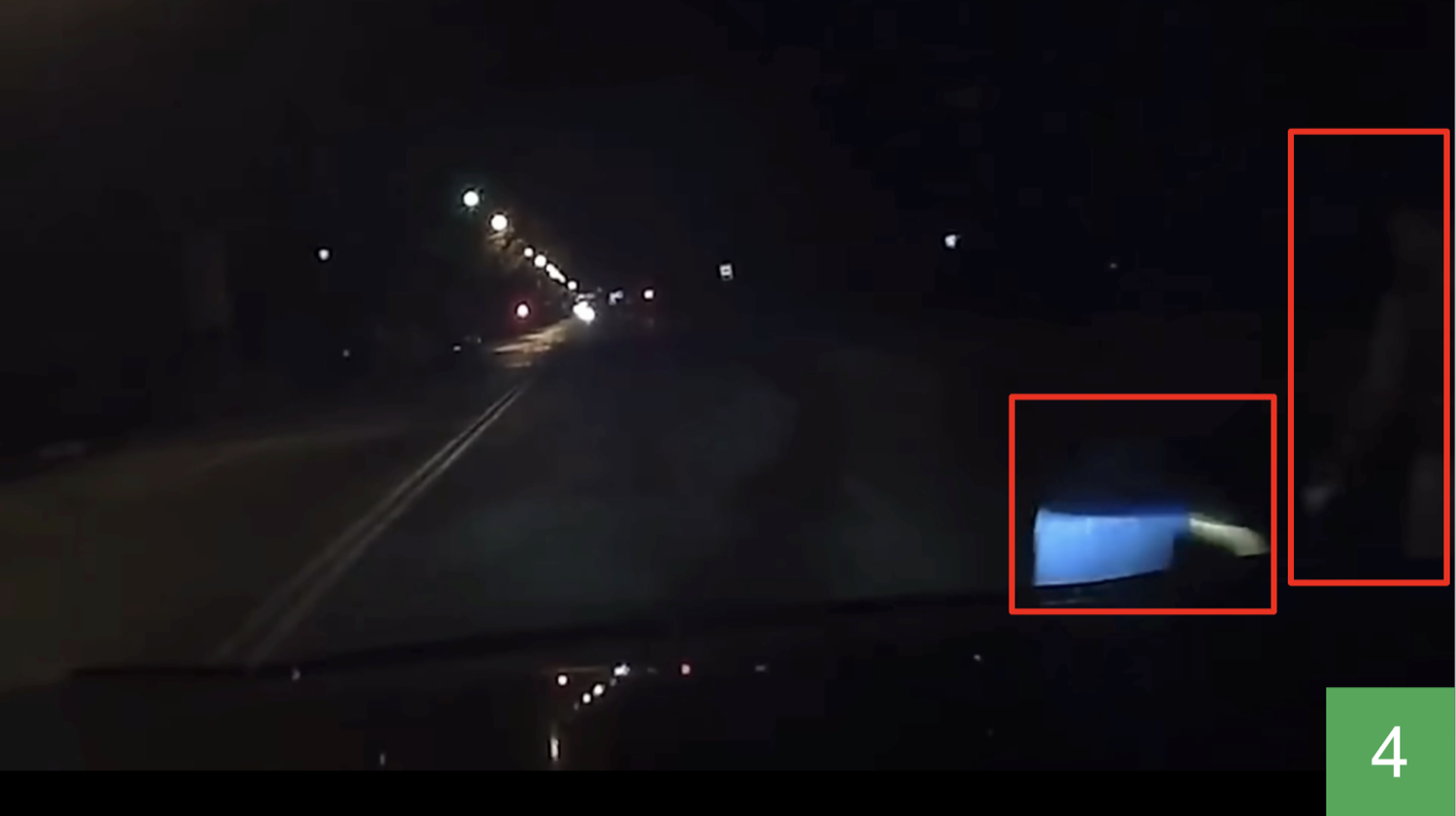}
      \caption{Friend helps them up}
      \label{fig:archetype-fraud-4}
    \end{subfigure}
    
    \caption{Pedestrian from \cite{evidence-the-fraud-jump} stages an accident with friend.}
    \label{fig:the-fraud}
\end{figure}

The Con Artist stages exaggerated collisions to claim insurance compensation or extort money out of the driver. Fortunately, because the con artist wants to minimize harm, they typically wait for an at rest or slow vehicle before throwing themselves onto the vehicle, either from a blind spot or the front/back.

\begin{table}[htbp]
    \centering
    \begin{tabular}{|p{0.2\textwidth}|p{0.2\textwidth}|}
        \hline
        \textbf{Essential Behaviors} & \textbf{Optional Behaviors} \\
        \hline
        \begin{itemize}
            \item Collision
            \item Looking
            \item Fall
            \item Not-Cross
            \item Run into traffic
            \item Cross without crosswalk
        \end{itemize} &
        \begin{itemize}
            \item Climbing onto carhood
            \item Thrown-back
            \item Ignore traffic
        \end{itemize}\\
        \hline
    \end{tabular}
    \caption{Con Artist Archetype}
    \label{table:con-artist-archetype-behaviors}
\end{table}

In figure \ref{fig:the-fraud}, the pedestrian stands in the path of the incoming vehicle. As the vehicle slows down, the pedestrian jumps towards it, ensuring a collision. Upon impact, the pedestrian pretends to fall onto the ground, exaggerating the impact. Another individual, also part of the act, quickly comes to help.


\subsection{The Foreigner}\label{sec:the-foreigner}

\begin{figure}[bt!]

    \begin{subfigure}[b]{.49\linewidth}
      \centering
      \includegraphics[width=.99\linewidth]{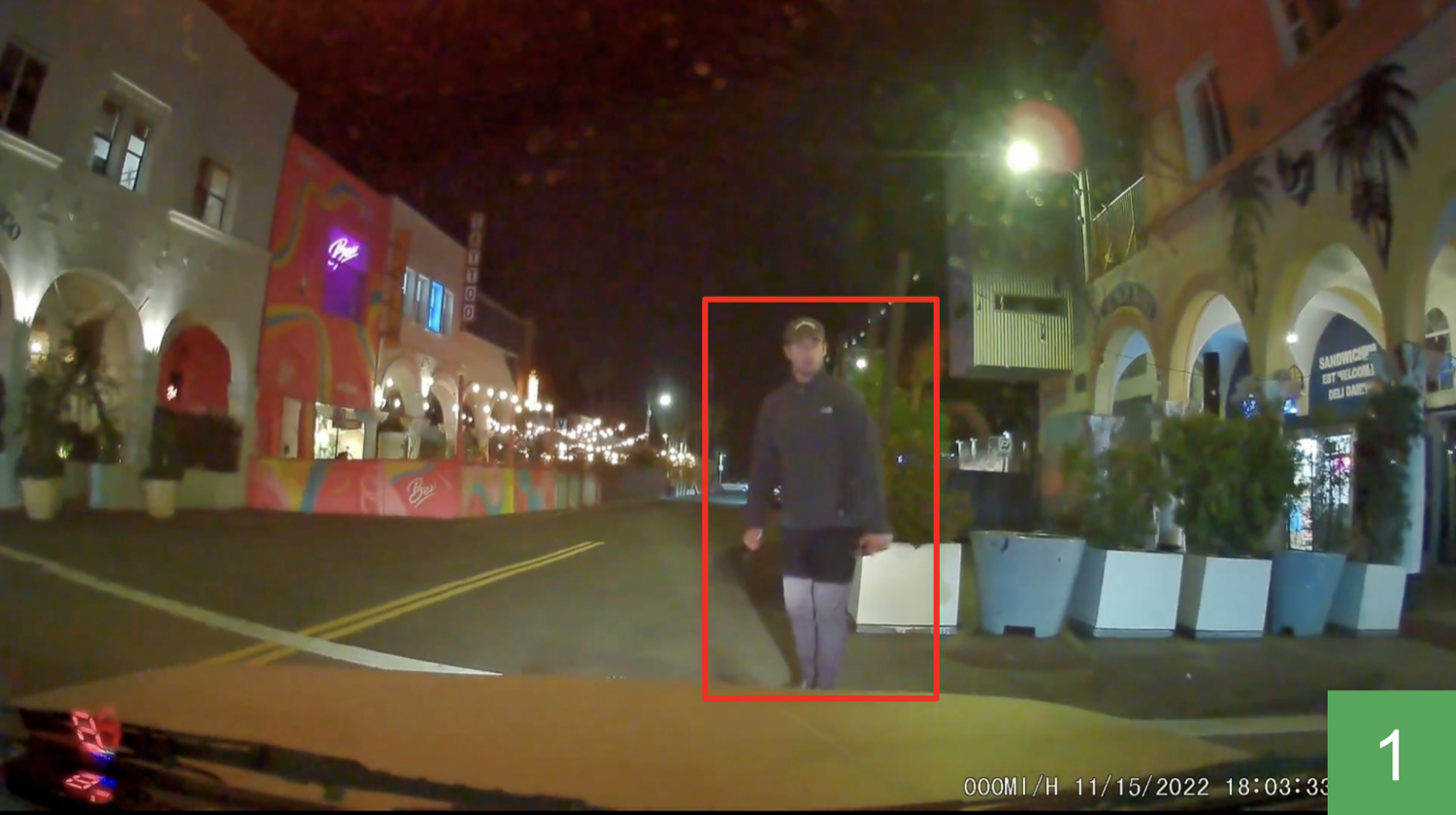}
      \caption{Confused by car approaching}
      \label{fig:archetype-tourist-1}
    \end{subfigure}
    \hfill
    \begin{subfigure}[b]{.49\linewidth}
      \centering
      \includegraphics[width=.99\linewidth]{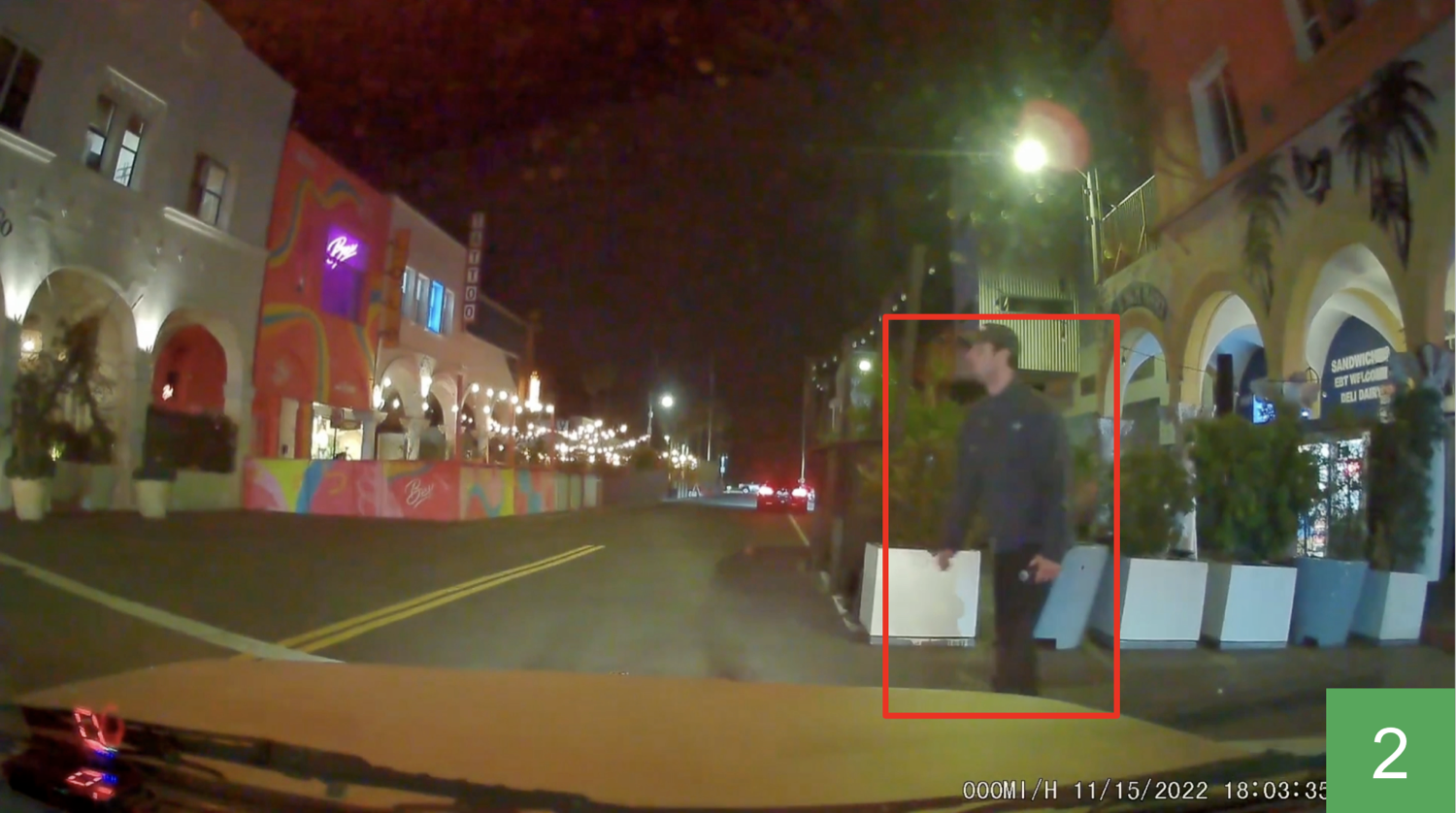}
      \caption{Looks around \& steps back}
      \label{fig:archetype-tourist-2}
    \end{subfigure}      
    \caption{Confused Venice tourist unsure what to do. \cite{evidence-the-confused-tourist-venice}}
    \label{fig:the-tourist}
\end{figure}

The Foreigner is a visitor unfamiliar with local traffic norms who misinterprets signals/structures, causing unsafe crossing decisions due to different traffic cultures. When traveling or immigrating to a new city, these pedestrians often lack awareness of local traffic laws, structure, and norms. 

\begin{table}[htbp]
    \centering
    \begin{tabular}{|p{0.2\textwidth}|p{0.2\textwidth}|}
        \hline
        \textbf{Essential Behaviors} & \textbf{Optional Behaviors} \\
        \hline
        \begin{itemize}
            \item Ignore traffic
            \item Near-miss
            \item Cross without crosswalk
            \item Run into traffic
            \item Not looking/glancing
        \end{itemize} &
        \begin{itemize}
            \item Retreat
            \item Back-turned
            \item Collision
        \end{itemize}\\
        \hline
    \end{tabular}
    \caption{Foreigner Archetype}
    \label{table:foreigner-archetype-behaviors}
\end{table}

In Venice, California, a pedestrian is confused by a 3-way intersection (see figure \ref{fig:the-tourist}). Video captions reveal a pedestrian walking mode, but most don't use it. Instead, the pedestrian attempts to cross and is met by the incoming vehicle. The puzzled pedestrian then looks around before retreating and allowing the vehicle to pass.


\subsection{The Influencer}\label{sec:the-influencer}

\begin{figure}[bt!]
    \centering
    \framebox{\includegraphics[width=.9\linewidth]{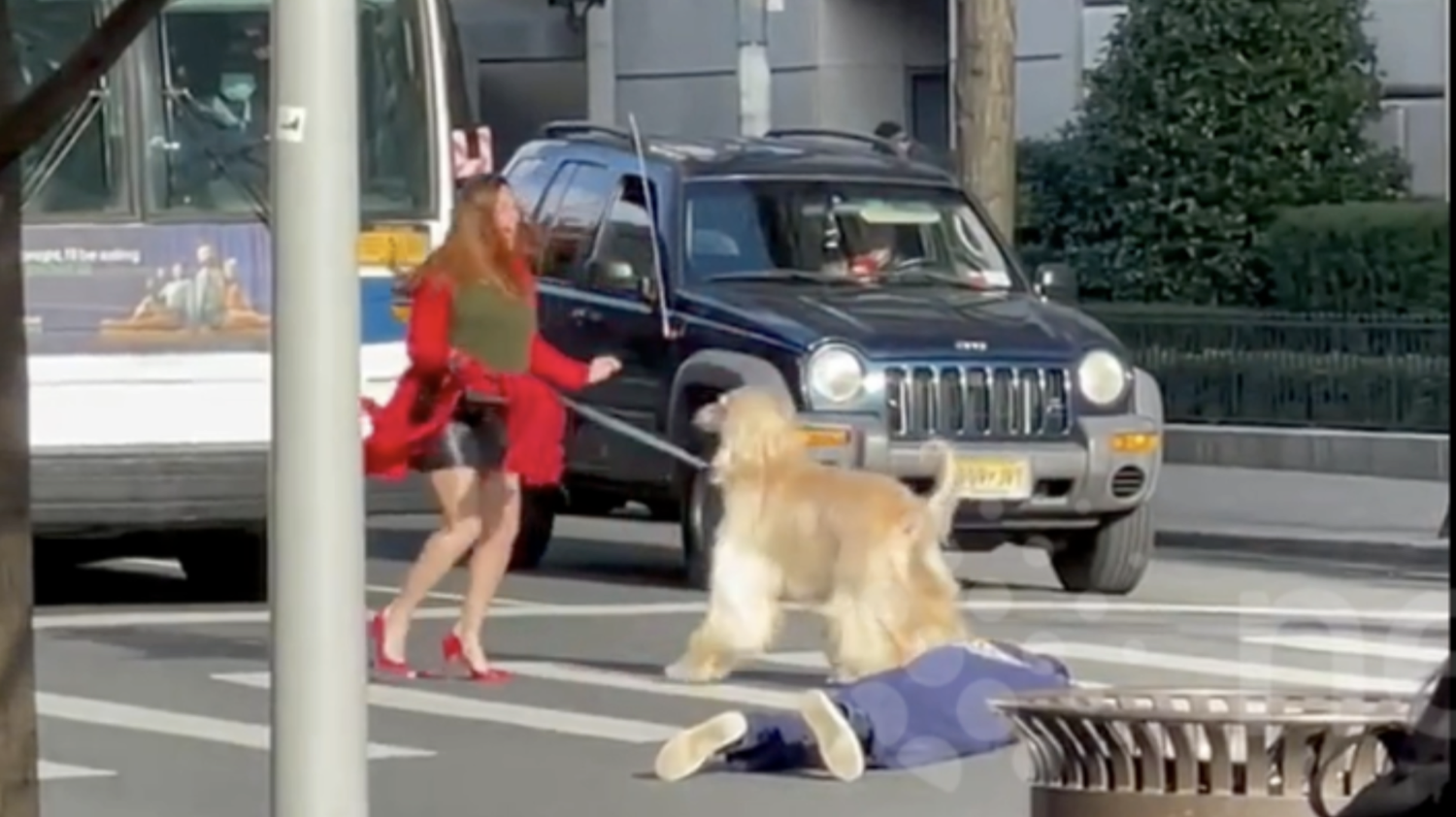}}
    \caption{Influencer with dog as photographer lies down \cite{evidence-the-influencer}.}
    \label{fig:archetype-influencer}
\end{figure}

The Influencer is a content creator who disregards traffic regulations to film videos, often involving photographers lying on roads or extended occupation of traffic lanes.

\begin{table}[htbp]
    \centering
    \begin{tabular}{|p{0.2\textwidth}|p{0.2\textwidth}|}
        \hline
        \textbf{Essential Behaviors} & \textbf{Optional Behaviors} \\
        \hline
        \begin{itemize}
            \item Ignore traffic
            \item Gesturing
            \item Glancing
        \end{itemize}
        & \\
        \hline
    \end{tabular}
    \caption{Influencer Archetype}
    \label{table:influencer-archetype-behaviors}
\end{table}

In figure \ref{fig:archetype-influencer}, the woman stops New York traffic to record Christmas dancing footage with her dog. The photographer is seen filming on the ground. Upon finishing, they quickly run back to the sidewalk. Tragedy would arise if the photographer goes undetected by an AV.

\subsection{The Protester}\label{sec:the-protester}

\begin{figure}[bt!]
    \centering
    \begin{subfigure}[b]{.49\linewidth}
      \centering
      \includegraphics[width=.99\linewidth]{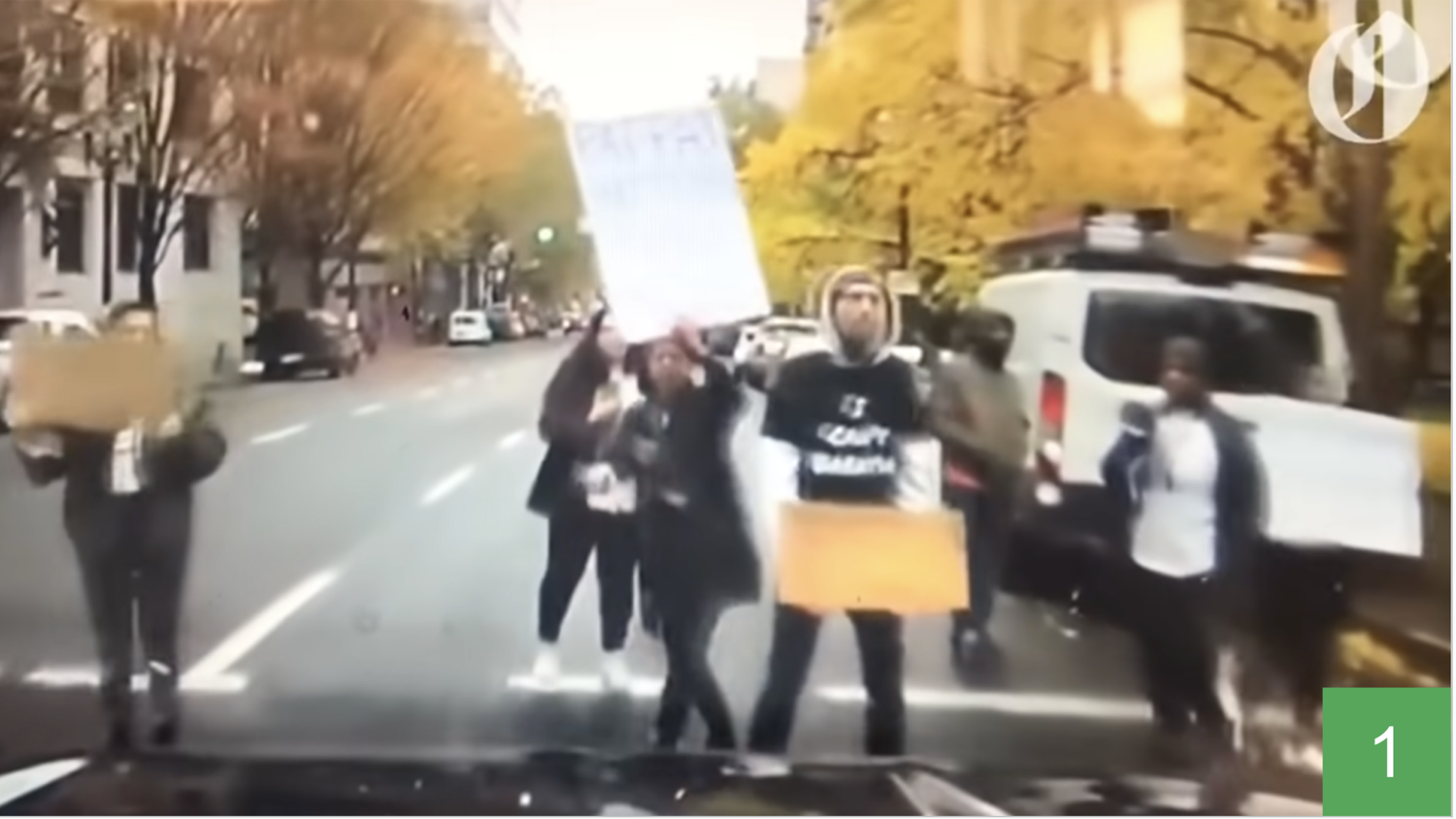}
      \caption{Protestors blocks road}
      \label{fig:archetype-protester-1}
    \end{subfigure}
    \hfill
    \begin{subfigure}[b]{.49\linewidth}
      \centering
      \includegraphics[width=.99\linewidth]{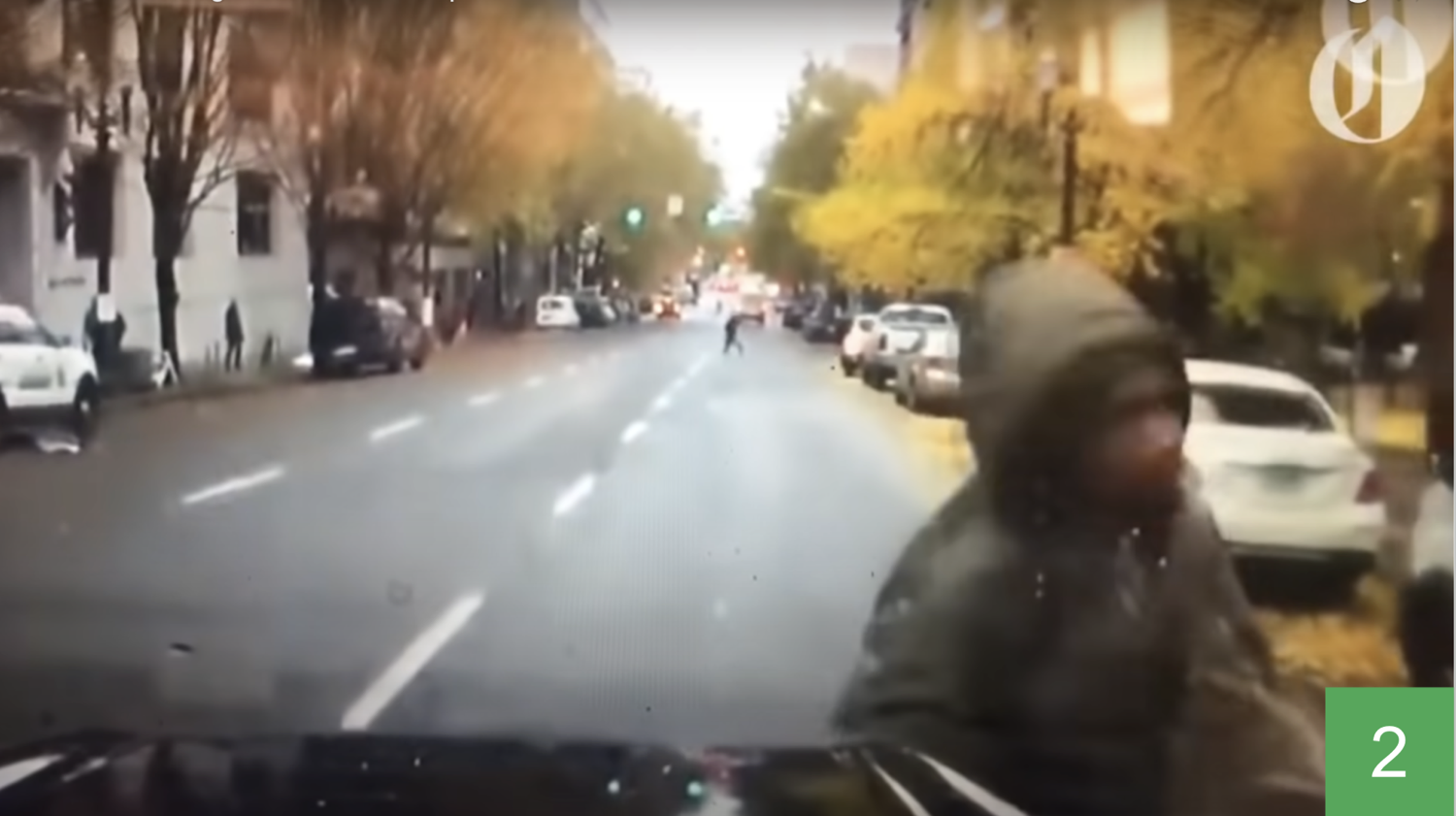}
      \caption{One stays in place \& is hit}
      \label{fig:archetype-protester-2}
    \end{subfigure}
    
    \begin{subfigure}[b]{.49\linewidth}
      \centering
      \includegraphics[width=.99\linewidth]{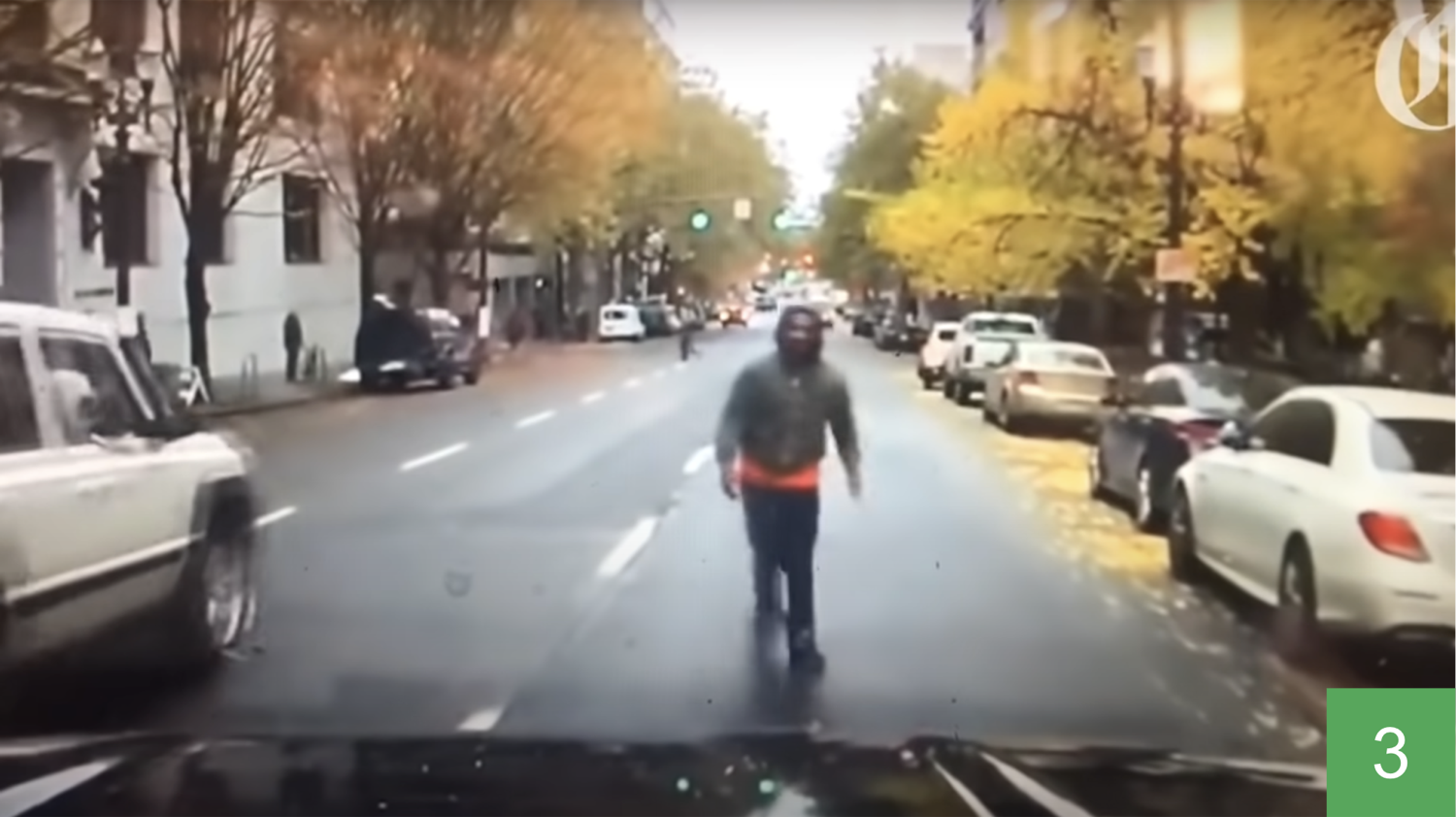}
      \caption{Knocked back, still in path}
      \label{fig:archetype-protester-3}
    \end{subfigure}
    \hfill
    \begin{subfigure}[b]{.49\linewidth}
      \centering
      \includegraphics[width=.99\linewidth]{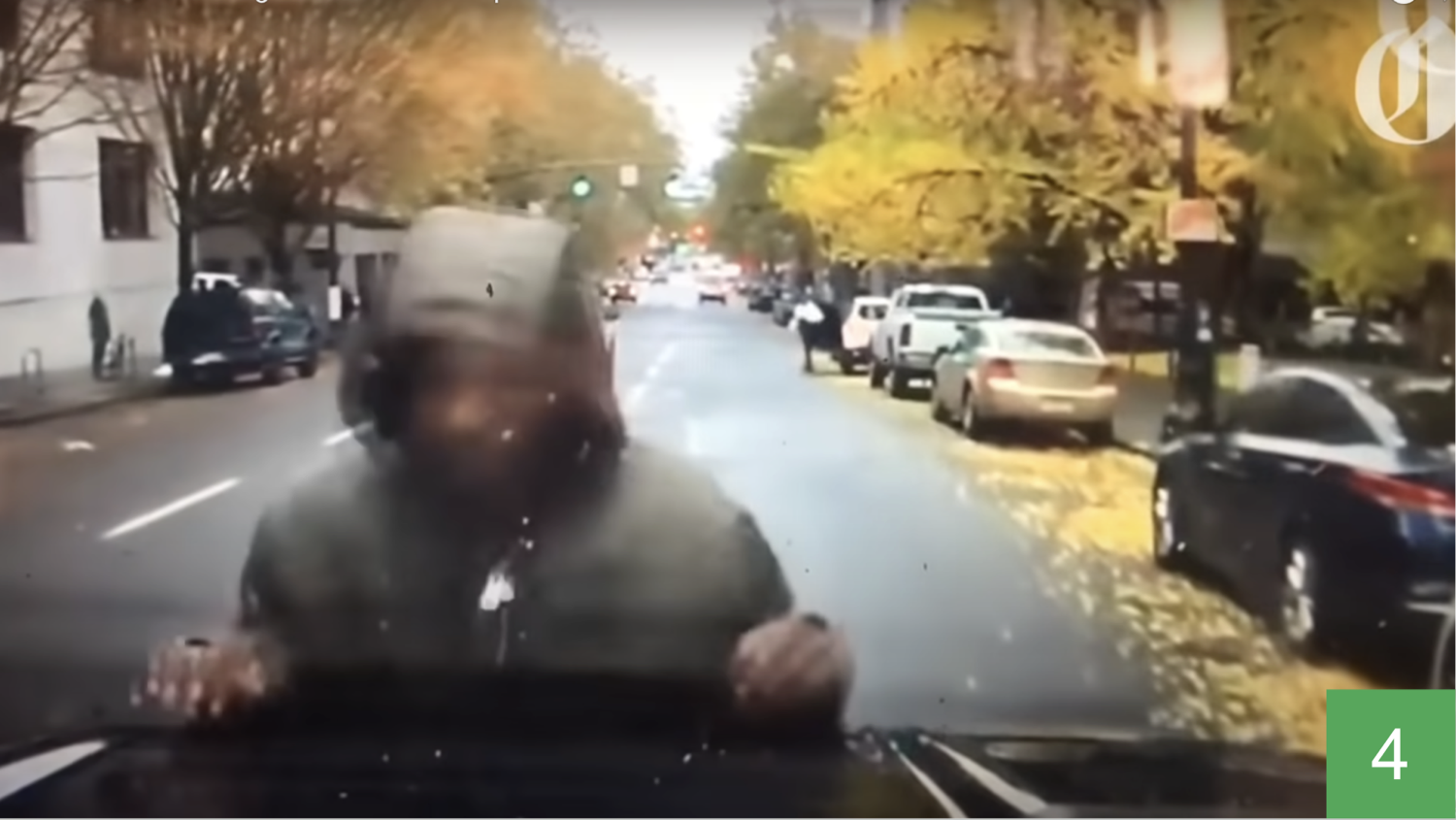}
      \caption{Vehicle continues \& hits again}
      \label{fig:archetype-protester-4}
    \end{subfigure}
    
    \caption{A group of pedestrians from \cite{evidence-the-protester} blocks the road as a vehicle passes through. One pedestrian is hit and remains in the path, causing a second collision.}
    \label{fig:the-protester}
\end{figure}

The Protester frequently ignores traffic regulations, occupying driving lanes. These actions can disrupt traffic and, in some cases, trigger aggressive responses from human drivers.

\begin{table}[htbp]
    \centering
    \begin{tabular}{|p{0.2\textwidth}|p{0.2\textwidth}|}
        \hline
        \textbf{Essential Behaviors} & \textbf{Optional Behaviors} \\
        \hline
        \begin{itemize}
            \item Frozen
            \item Not-cross
            \item Looking
            \item Ignore-traffic
            \item Agitated
        \end{itemize} &
        \begin{itemize}
            \item Group-walk
            \item Along-lane
            \item Back-turned
        \end{itemize}\\
        \hline
    \end{tabular}
    \caption{Protester Archetype}
    \label{table:protester-archetype-behaviors}
\end{table}

In figure \ref{fig:the-protester}, a driver deliberately strikes a protester with his truck.

\subsection{The Confronted}\label{sec:the-confronted}

\begin{figure}[bt!]
    \centering
    \begin{subfigure}[b]{.49\linewidth}
      \centering
      \includegraphics[width=.99\linewidth]{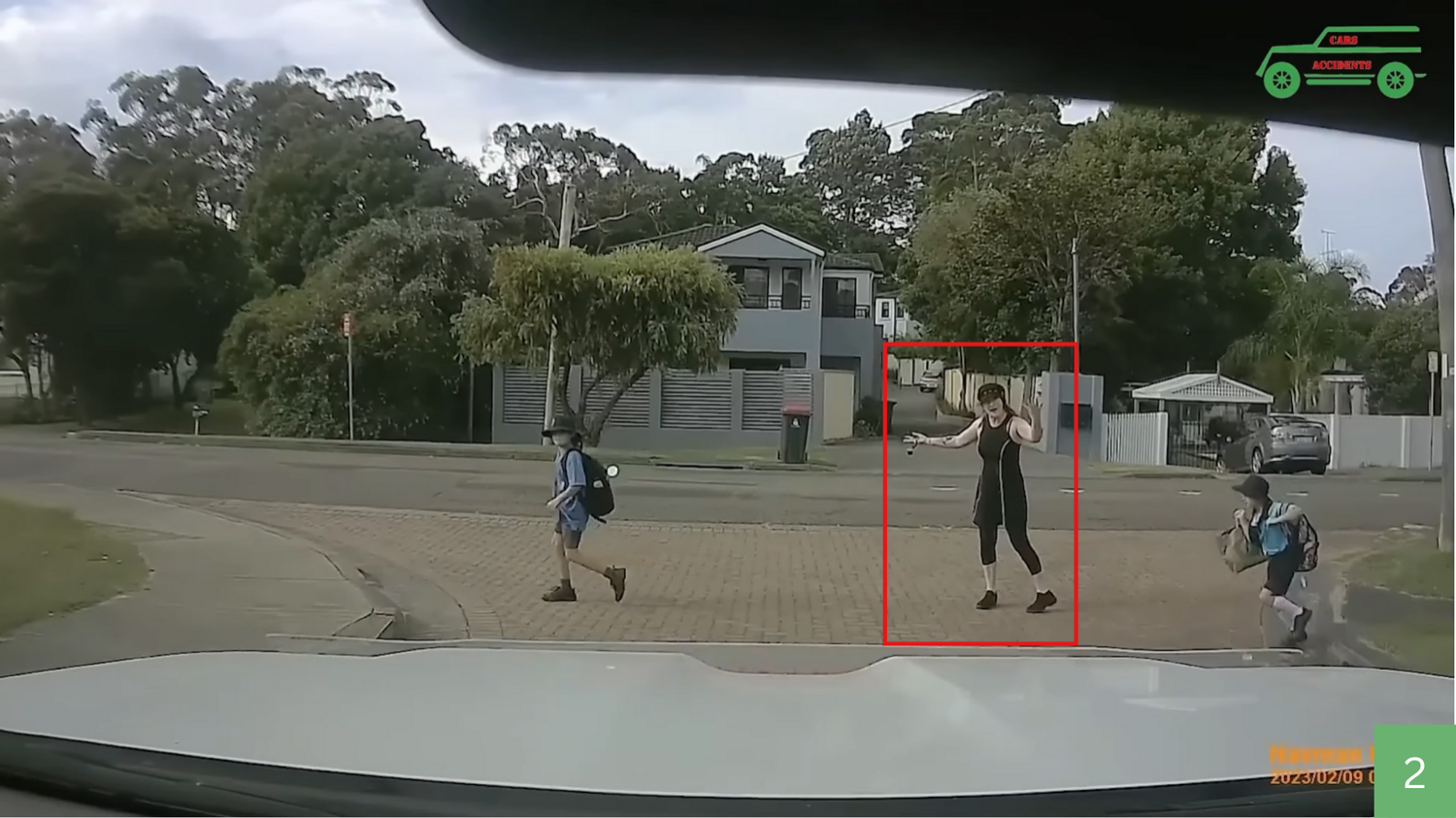}
      \caption{Begins aggressive gestures.}
      \label{fig:archetype-confronted-1}
    \end{subfigure}
    \hfill
    \begin{subfigure}[b]{.49\linewidth}
      \centering
      \includegraphics[width=.99\linewidth]{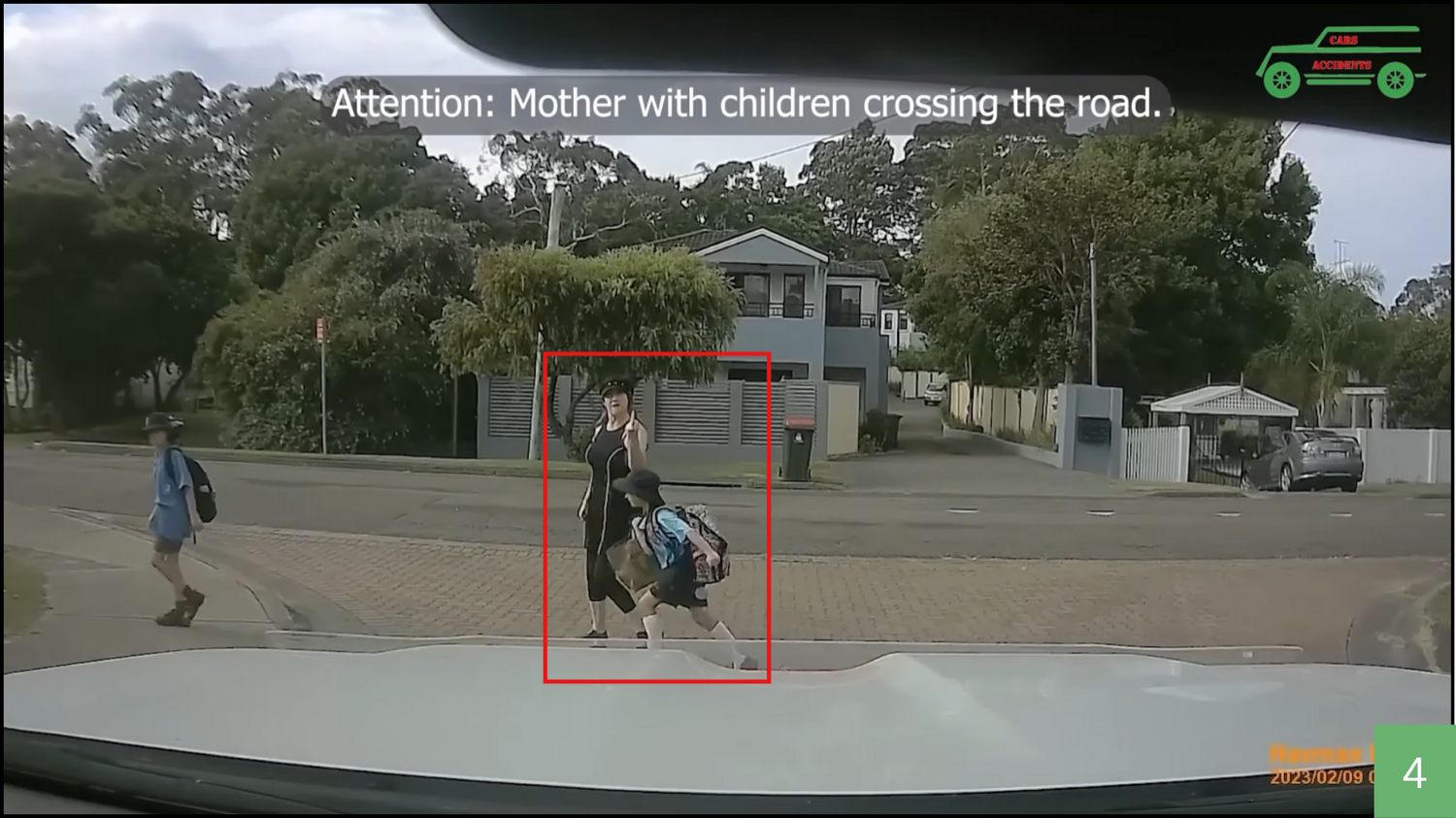}
      \caption{Puts up middle finger.}
      \label{fig:archetype-confronted-2}
    \end{subfigure}
    
    
    \caption{A hostile mother \cite{evidence-the-confronted} as she crosses with children.}
    \label{fig:the-confronted}
\end{figure}

The Confronted's behavior is driven by hostile and emotional responses. When interacting with drivers, they often show visible agitation, confrontational gestures, or even attack the vehicle, regardless of who is at fault.

\begin{table}[htbp]
    \centering
    \begin{tabular}{|p{0.2\textwidth}|p{0.2\textwidth}|}
        \hline
        \textbf{Essential Behaviors} & \textbf{Optional Behaviors} \\
        \hline
        \begin{itemize}
            \item Agitated
            \item Aggression
            \item Cross
            \item Ignore traffic
        \end{itemize} &
        \begin{itemize}
            \item Assault
            \item Gesturing
            \item Near-miss
        \end{itemize}\\
        \hline
    \end{tabular}
    \caption{Confronted Archetype}
    \label{table:confronted-archetype-behaviors}
\end{table}

In figure \ref{fig:the-confronted}, a mother crosses with her 2 children. The mother pauses several times to gesture and express her emotions towards the vehicle.

\subsection{The Pseudo Pedestrian}\label{sec:the-pseudo-pedestrian}

\begin{figure}[bt!]
    \centering
    \begin{subfigure}[b]{.49\linewidth}
      \centering
      \includegraphics[width=.99\linewidth]{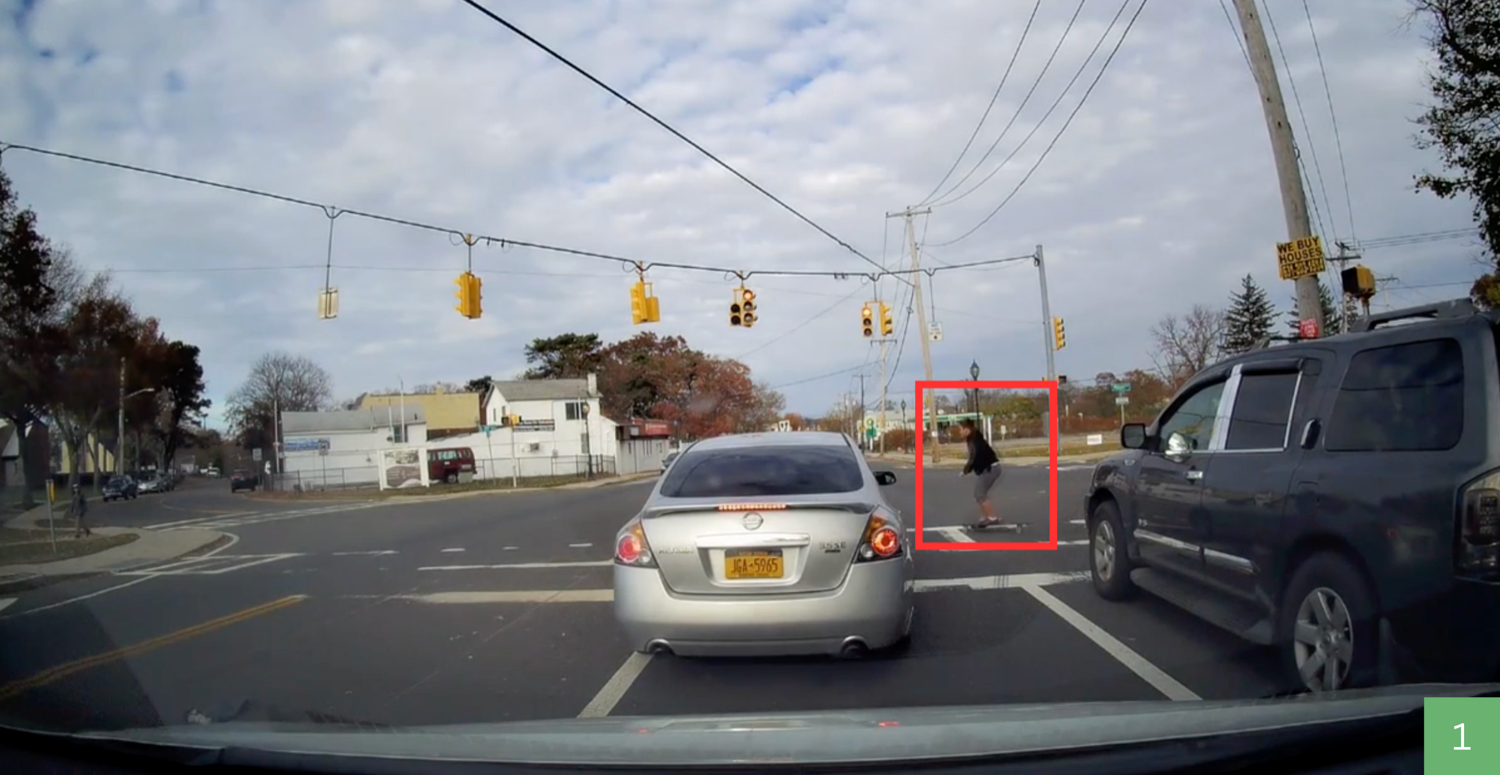}
      \caption{Crosses fast with skateboard.}
      \label{fig:archetype-pseudopedestrian-1}
    \end{subfigure}
    \hfill
    \begin{subfigure}[b]{.49\linewidth}
      \centering
      \includegraphics[width=.99\linewidth]{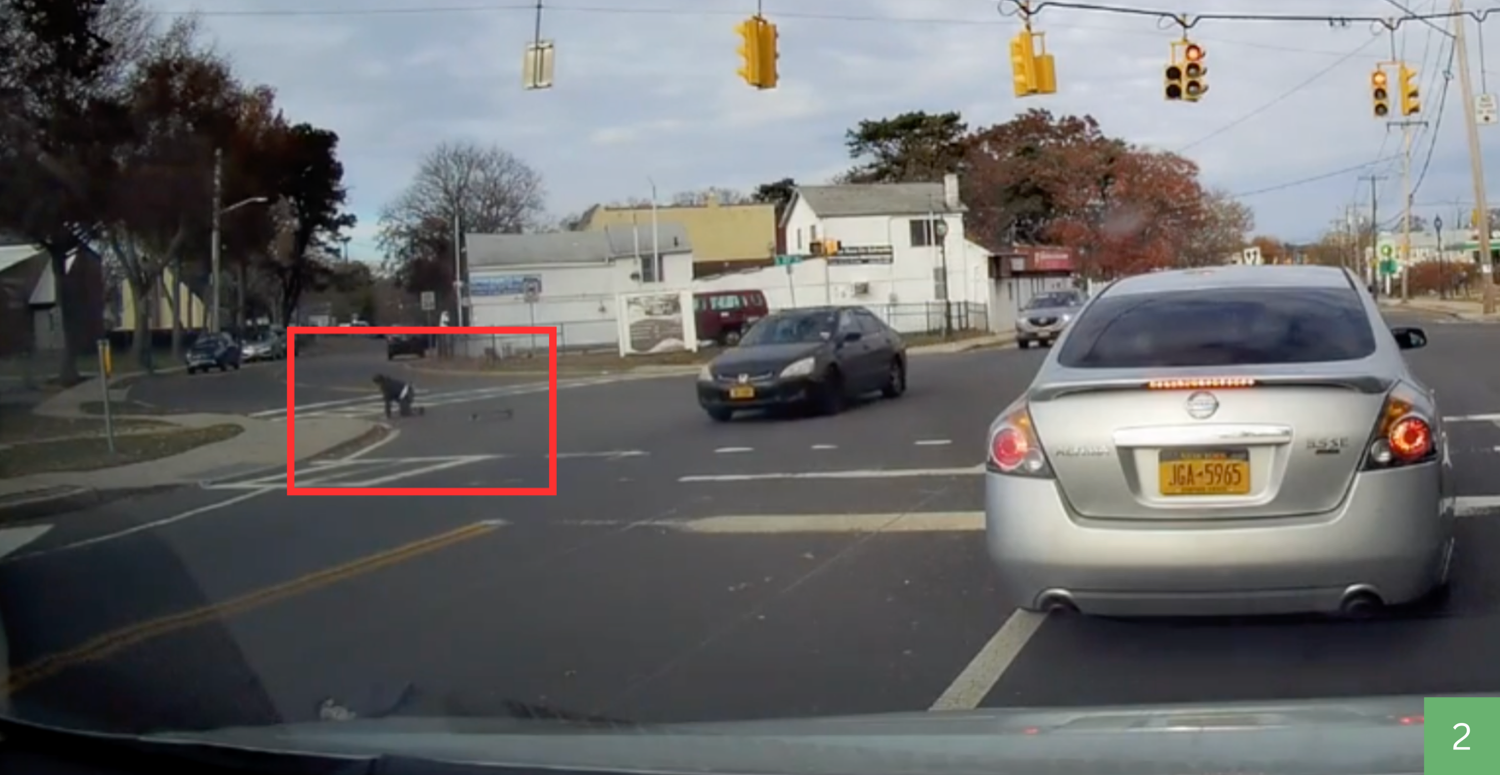}
      \caption{Falls, skateboard rolls off.}
      \label{fig:archetype-pseudopedestrian-2}
    \end{subfigure}
    
    \begin{subfigure}[b]{.49\linewidth}
      \centering
      \includegraphics[width=.99\linewidth]{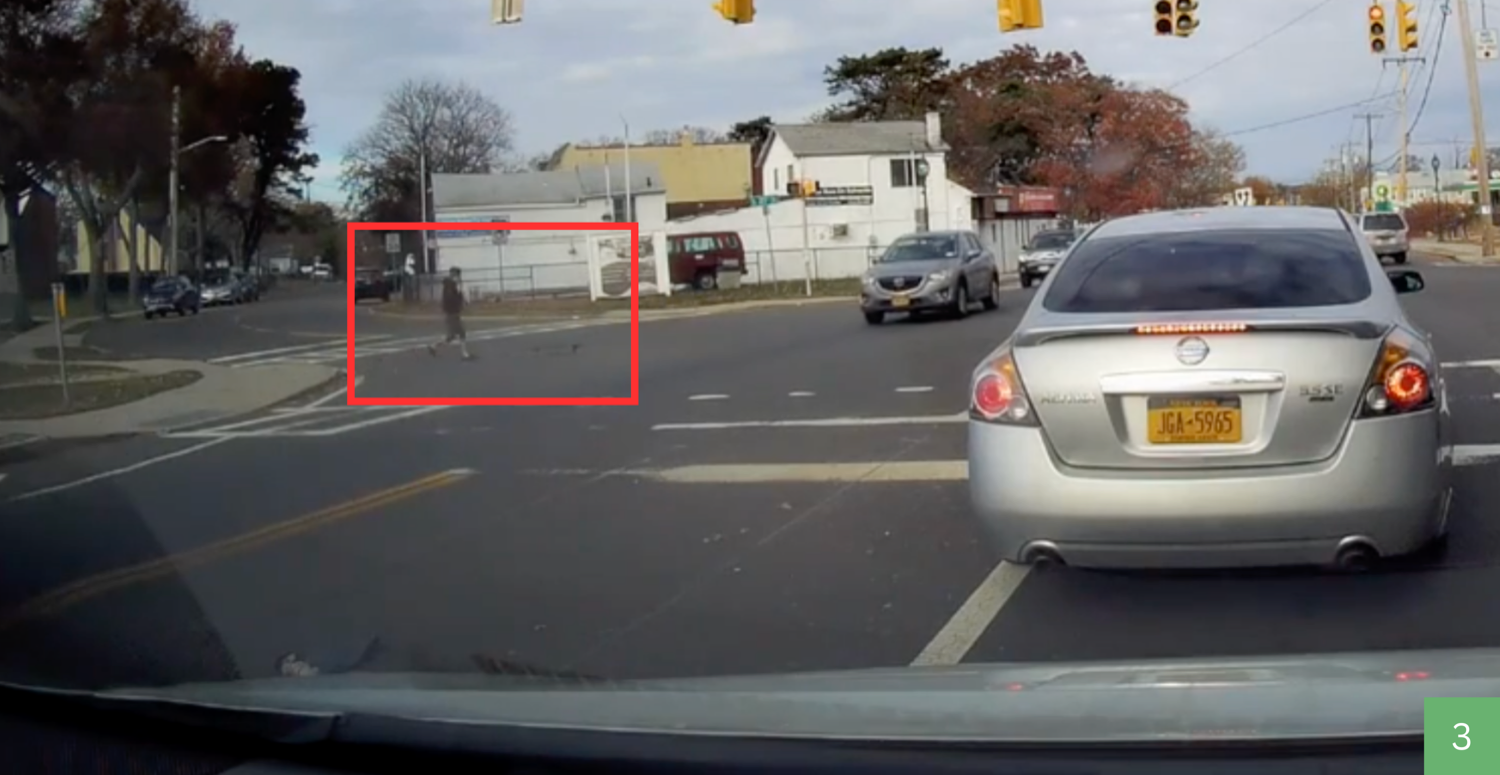}
      \caption{Runs after skateboard.}
      \label{fig:archetype-pseudopedestrian-3}
    \end{subfigure}
    \hfill
    \begin{subfigure}[b]{.49\linewidth}
      \centering
      \includegraphics[width=.99\linewidth]{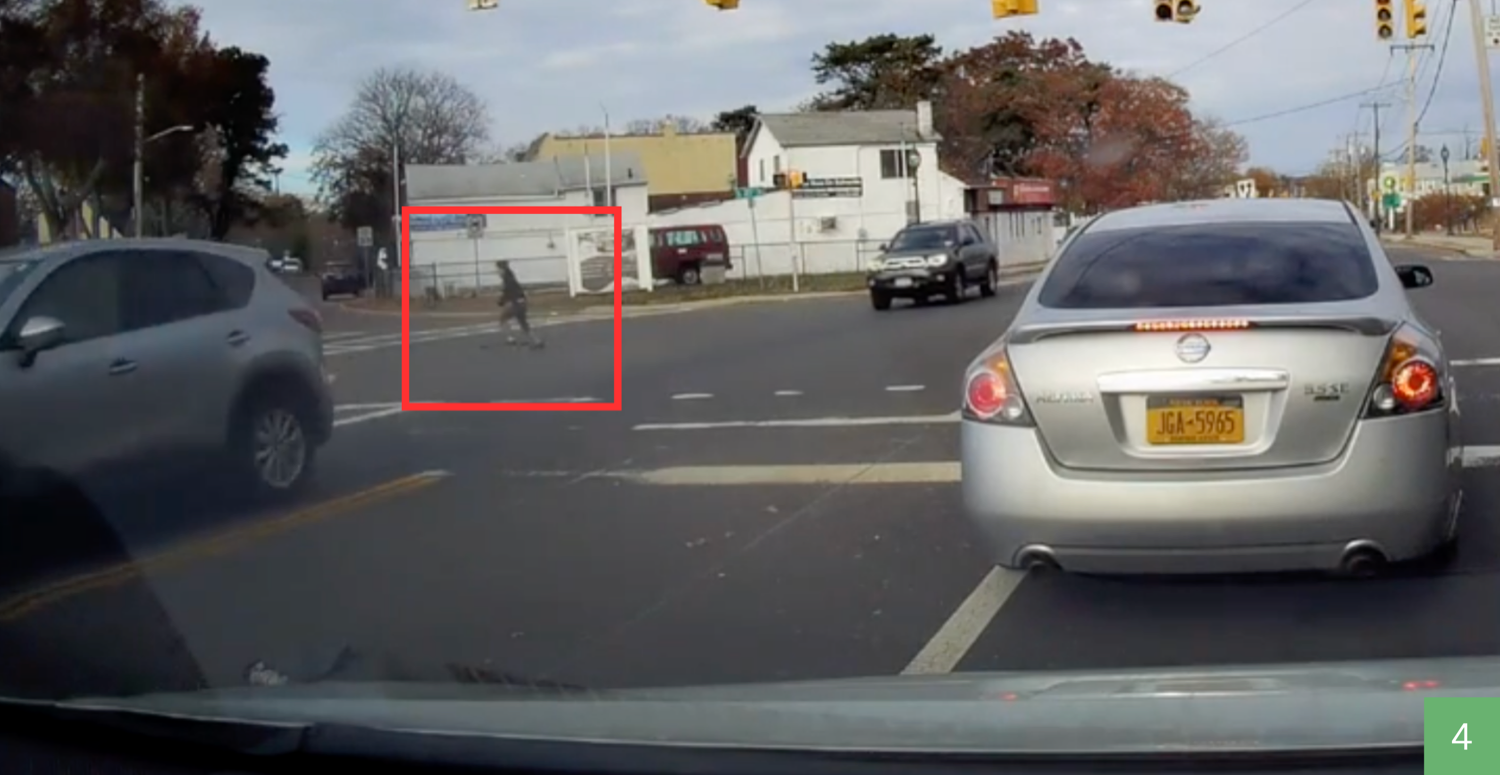}
      \caption{Finally skates away.}
      \label{fig:archetype-pseudopedestrian-4}
    \end{subfigure}
    
    \caption{Skateboarder from \cite{evidence-the-pseudo-pedestrian} loses control and falls off, pushing skateboard into traffic and running after it.}
    \label{fig:the-pseudopedestrian}
\end{figure}

The Pseudo Pedestrian moves fast and unpredictably via wheels. They are found rollerblading, skateboarding, or in wheelchairs.

\begin{table}[htbp]
    \centering
    \begin{tabular}{|p{0.2\textwidth}|p{0.2\textwidth}|}
        \hline
        \textbf{Essential Behaviors} & \textbf{Optional Behaviors} \\
        \hline
        \begin{itemize}
            \item Ignore traffic
            \item Run into traffic
            \item Cross
            \item Collision
            \item Not looking/glancing
        \end{itemize} &
        \begin{itemize}
            \item Along-lane
            \item Swerve
            \item Pop-out-occlusion
        \end{itemize}\\
        \hline
    \end{tabular}
    \caption{Pseudo Pedestrians Archetype}
    \label{table:pseudo-pedestrians-archetype-behaviors}
\end{table}

In \ref{fig:the-pseudopedestrian}, a pedestrian skateboards across the road, accelerating. However, they hit the sidewalk, lose balance, and fall. Their skateboard rolls into traffic, and the skateboarder chases after it. Following retrieval, the pedestrian skates away.

\subsection{The Street Vendor}\label{sec:the-street-vendor}

\begin{figure}[bt!]
    \centering
    \begin{subfigure}[b]{.49\linewidth}
      \centering
      \includegraphics[width=.99\linewidth]{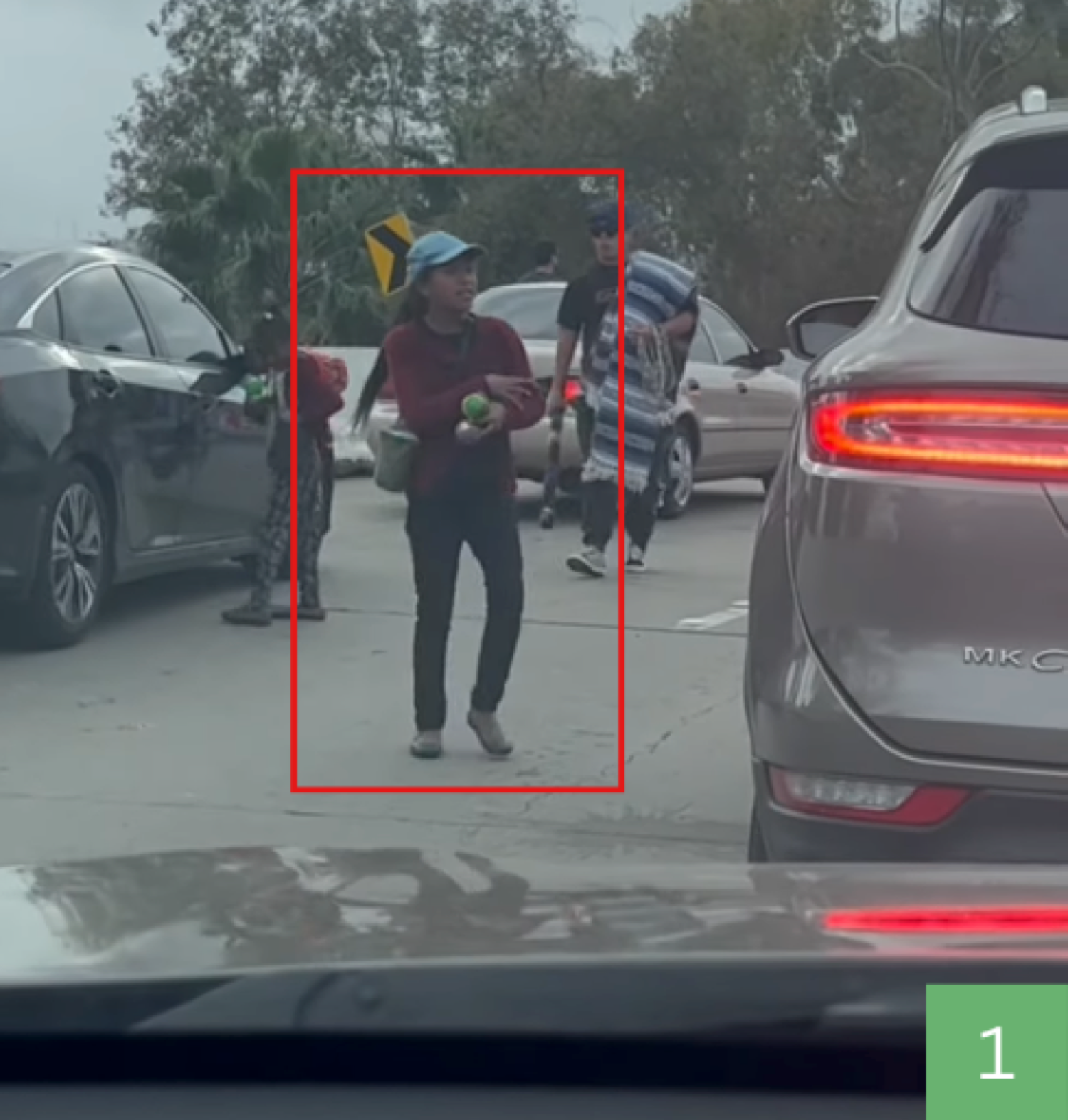}
      \caption{Looks at \& walks towards car.}
      \label{fig:archetype-streetvendor-1}
    \end{subfigure}
    \hfill
    \begin{subfigure}[b]{.49\linewidth}
      \centering
      \includegraphics[width=.99\linewidth]{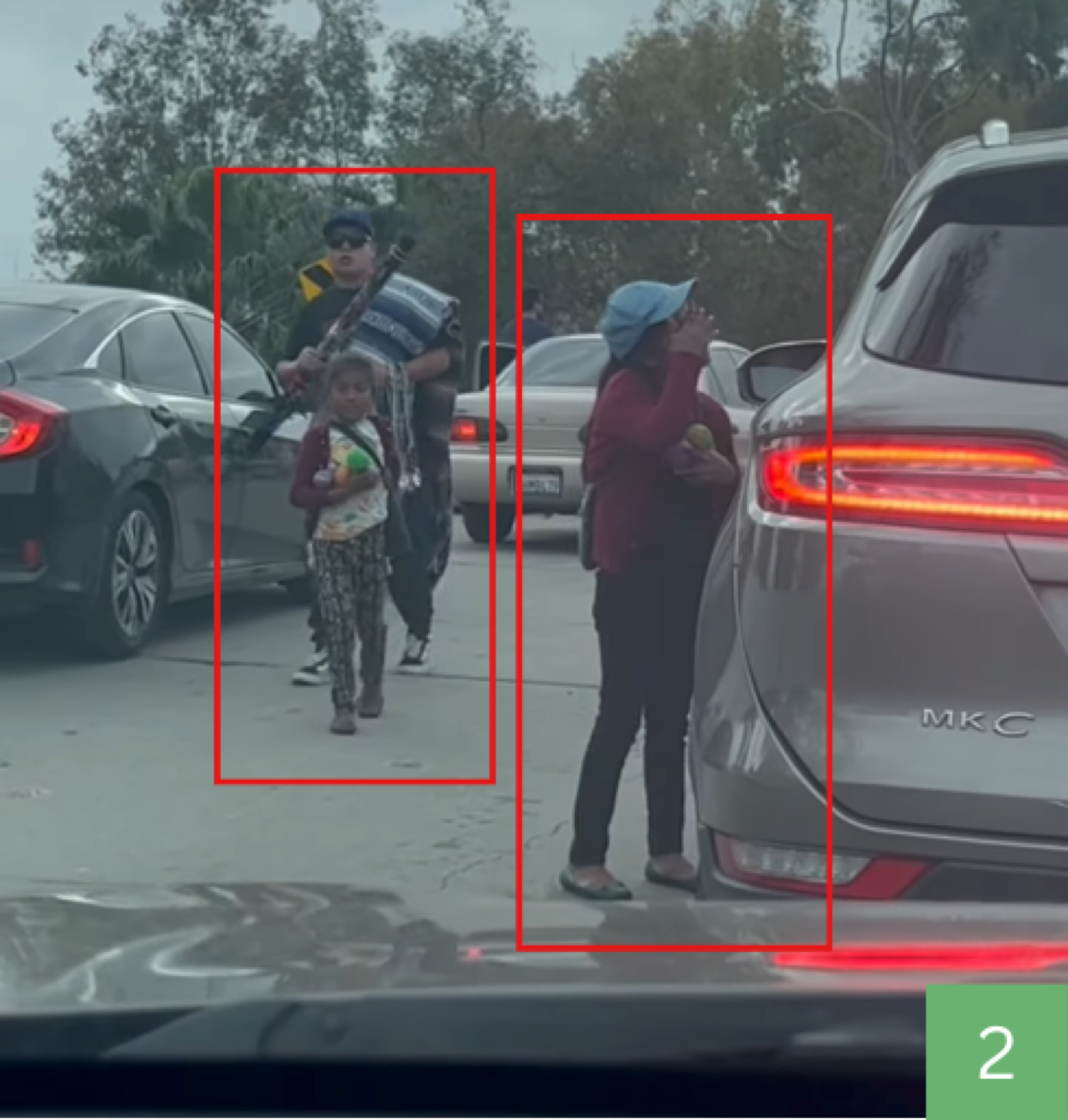}
      \caption{Vendor stops next to car.}
      \label{fig:archetype-streetvendor-2}
    \end{subfigure}

    \caption{Street vendors \cite{evidence-the-street-vendor} walk along slow-moving traffic.}
    \label{fig:the-streetvendor}
\end{figure}

The Street Vendor approaches cars during slow or stopped traffic to market their products, made risky by their close proximity with vehicles.

\begin{table}[htbp]
    \centering
    \begin{tabular}{|p{0.2\textwidth}|p{0.2\textwidth}|}
        \hline
        \textbf{Essential Behaviors} & \textbf{Optional Behaviors} \\
        \hline
        \begin{itemize}
            \item Pause-start
            \item Pickup-object
            \item Cross-on-red
        \end{itemize}
        & \\
        \hline
    \end{tabular}
    \caption{Street Vendor Archetype}
    \label{table:street-vendor-archetype-behaviors}
\end{table}

In figure \ref{fig:the-streetvendor}, street vendors walk alongside the moving cars on the left. On the right, they can also be seen approaching the driver's side with merchandise in hand.

\section{Conclusion}\label{conclusion}
This preprint extends our pedestrian archetype framework by introducing 7 new archetypes identified through further YouTube dash-cam video annotation. These archetypes capture observable pedestrian behavior patterns that are meaningfully different from the 12 archetypes proposed in our prior work.

Our main argument remains the same: dangerous pedestrian behavior should not be modeled only as isolated behavior tags. Pedestrians often show connected patterns of behavior, and these patterns matter for autonomous vehicle safety testing. Archetypes help organize these patterns into communicable pedestrian models that can be used for annotation, simulation, and evaluation.

By expanding the archetype taxonomy, we move closer to a more complete behavioral test space for autonomous vehicles. The long-term goal is to help AV systems prepare not only for law-abiding pedestrians, but also for the rare, dangerous, and unpredictable pedestrians that matter most for safety.

\appendix
\begin{enumerate}
    \item \href{https://pedanalyze.readthedocs.io/en/latest/pedestrian-behavior-tags.html}{https://pedanalyze.readthedocs.io/en/latest/pedestrian-behavior-tags.html}\label{appendix-pedestrian-behaviors}
\end{enumerate}









\bibliographystyle{IEEEtran}
\bibliography{IEEEabrv,references, references-behavior-evidence}

\end{document}